\documentclass[10pt,twocolumn,letterpaper]{article}

\usepackage{iccv}
\usepackage{times}
\usepackage{epsfig}
\usepackage{graphicx}
\usepackage{amsmath}
\usepackage{amssymb}
\usepackage{subfigure}
\usepackage{booktabs}
\usepackage{multirow}
\usepackage[ruled]{algorithm2e}

\usepackage[pagebackref=true,breaklinks=true,letterpaper=true,colorlinks,bookmarks=false]{hyperref}

\iccvfinalcopy % *** Uncomment this line for the final submission

\def\iccvPaperID{6618} % *** Enter the ICCV Paper ID here

\ificcvfinal\fi

\begin{document}

%%%%%%%%% TITLE
% Object Detection and Instance Segmentation}
% \title{Enhancing Feature Aggregation for Video Object Detection by Blending Temporal Relations and Local Representatives}
\title{TF-Blender: Temporal Feature Blender for Video Object Detection}

\author{Yiming Cui$^{1}$\thanks{Equal contributions.}, Liqi Yan$^{2*}$, Zhiwen Cao$^{3}$, Dongfang Liu$^{4}$\thanks{Corresponding author.}\\
$^{1}$University of Florida, USA  \\
$^{2}$Fudan University, China\\
$^{3}$Purdue University, USA  \\
$^{4}$Rochester Institute of Technology, USA\\
{\tt\small cuiyiming@ufl.edu, \tt\small   yanliqi@westlake.edu.cn, \tt\small   cao270@purdue.edu,  \tt\small dongfang.liu@rit.edu}
}

% \author{Yiming Cui\\
% University of Florida\\
% {\tt\small cuiyiming@ufl.edu}
% \and
% Liqi Yan\\
% Fudan University\\
% % First line of institution2 address\\
% {\tt\small yanliqi@westlake.edu.cn}

% \and
% Zhiwen Cao\\
% Purdue University\\
% % First line of institution2 address\\
% {\tt\small cao270@purdue.edu}

% \and
% Dongfang Liu\\
% Rochester Institute of Technology\\
% % First line of institution2 address\\
% {\tt\small dongfang.liu@rit.edu}
% }

\maketitle
\ificcvfinal\thispagestyle{empty}\fi

%%%%%%%%% ABSTRACT
\begin{abstract}
Video objection detection is a challenging task because isolated video frames may encounter appearance deterioration, which introduces great confusion for detection.
One of the popular solutions is to exploit the temporal information and enhance per-frame representation through aggregating features from neighboring frames. 
% Desipiting achieving improvements in detection,
% existing methods primarily focus on which features are used for aggregation rather than how to aggregate features. Specifically, these methods use global weights for each feature map during the aggregation process. 
% \textcolor{red}{Despite achieving improvements in detection,
% existing methods primarily focus on which features are used for aggregation rather than how to aggregate features. Specifically, these methods use global weights for every feature map which ignore the spatial information during the aggregation process. }
% In this paper, we propose an associative feature blender module to apply local weights so that feature maps are weighted averaged pixel by pixel during aggregation process.
% \textcolor{red}{In this paper, we propose a framework named Blender-Net to enhance feature aggregation for video object detection. Our Blender-Net contains three modules: 1) Temporal relation module explores the relations between the current frame and its neighboring frames across temporal domain and generates local weights for every neighboring feature map to preserve spatial information during feature aggregation process. 2). Feature adjustment module is used to improve the representation of every neighboring feature map before aggregation. 3) Feature blender module first enhances the results from temporal relation module and normalize those from feature adjustment module and then combine these results to generate the final aggregated features for the current frame.}
Despite achieving improvements in detection, existing methods focus on the selection of higher-level video frames for aggregation rather than modeling lower-level temporal relations to increase the feature representation.
% To address this limitation, we propose a framework named Blender-Net to enhance feature aggregation for video object detection. Our Blender-Net contains three modules:
To address this limitation, we propose a novel solution named TF-Blender, which includes three modules: 1) Temporal relation models the relations between the current frame and its neighboring frames to preserve spatial information. 2). Feature adjustment enriches the representation of every neighboring feature map; 3) Feature blender combines outputs from the first two modules and produces stronger features for the later detection tasks. 
% The three modules work collaboratively for the feature aggregation which benefits the detection behavior. 
For its simplicity, TF-Blender can be effortlessly plugged into any detection network to improve detection behavior. Extensive evaluations on ImageNet VID and YouTube-VIS benchmarks  indicate the performance guarantees of using TF-Blender on recent state-of-the-art methods. Code is available at  https://github.com/goodproj13/TF-Blender.
%   Video recognition is still a challenging task because of issues caused by motion blur, defocus, and so on. To solve these issues and make full use of the temporal information across frames, many methods introduce feature aggregation to improve the recognition results. Even though many of them get a good performance, they pay more attention to which features are used for aggregation rather than how to aggregate features. Therefore, these methods use global weights for each feature map during aggregation process. In this paper, we propose a associative feature blender module to apply local weights so that feature maps are weighted averaged pixel by pixel during aggregation process. Based on these two modules, our methods can help most of the state-of-the-arts methods improve their detection accuracy by a large margin on ImageNet VID benchmark and YouTube-VIS. Code will be available.
\end{abstract}

%%%%%%%%% BODY TEXT
\section{Introduction}
% The advances in deep learning have boosted the accuracy of image recognition in the domain of image classification \cite{he2016deep,huang2017densely,krizhevsky2012imagenet,simonyan2014very}, instance segmentation \cite{bolya2019yolact,chen2020blendmask,he2017mask,wang2020solo}, and object detection \cite{ren2015faster,redmon2018yolov3,tian2019fcos}. 
With the progress of learning-based computer vision, recent research efforts have been extended from image tasks to the more challenging video domains. Video tasks, such as object detection \cite{deng2009imagenet}, video instance segmentation ~\cite{yang2019video}, and multi-object tracking and segmentation \cite{voigtlaender2019mots}, hold valuable potentials for real-world applications \cite{9206716, voigtlaender2019mots,liu2020video, liu2020visual} (i.e., autonomous driving or video surveillance ).
\begin{figure}[!bt]
    \centering
    \subfigure[Visualization of feature aggregation process]{
    \includegraphics[width=8cm]{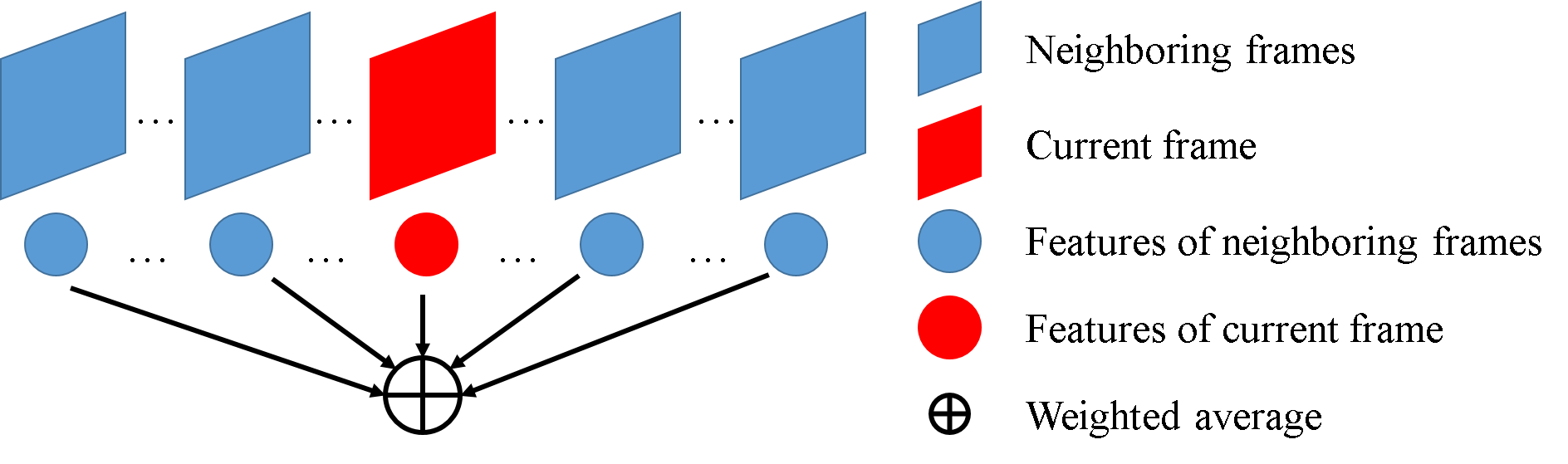}
    }
    \subfigure[Current aggregation methods]{
    \includegraphics[width=3cm]{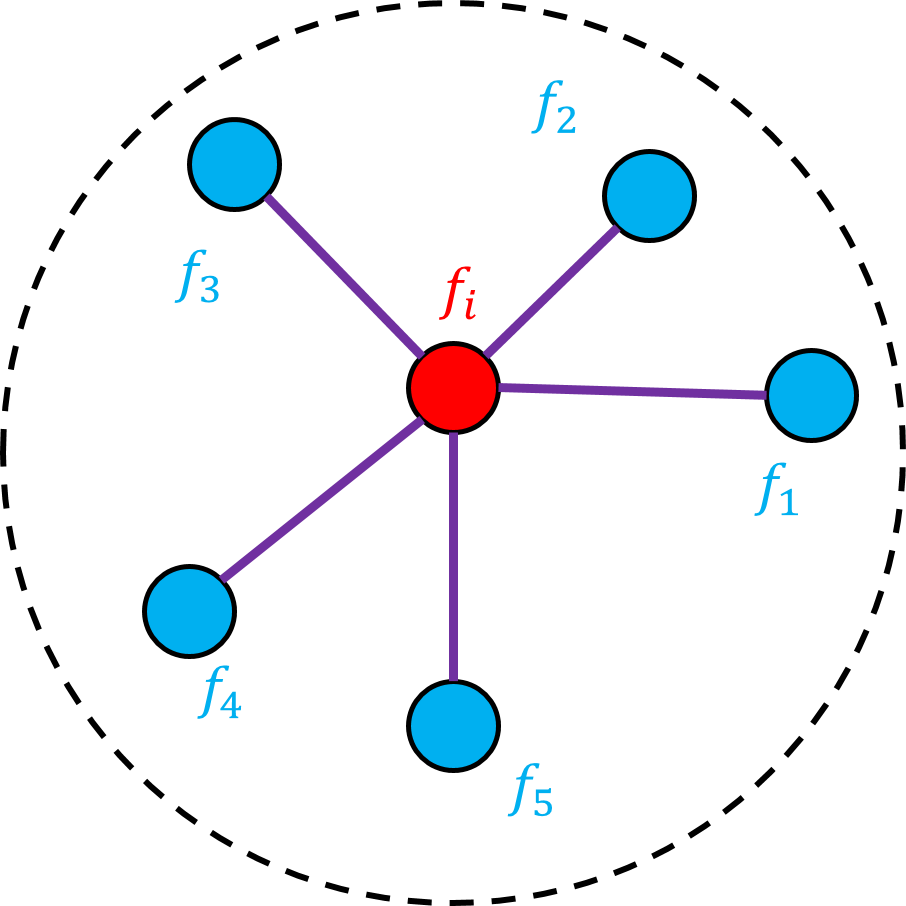}
    }
    \subfigure[Our aggregation methods]{
    \includegraphics[width=3cm]{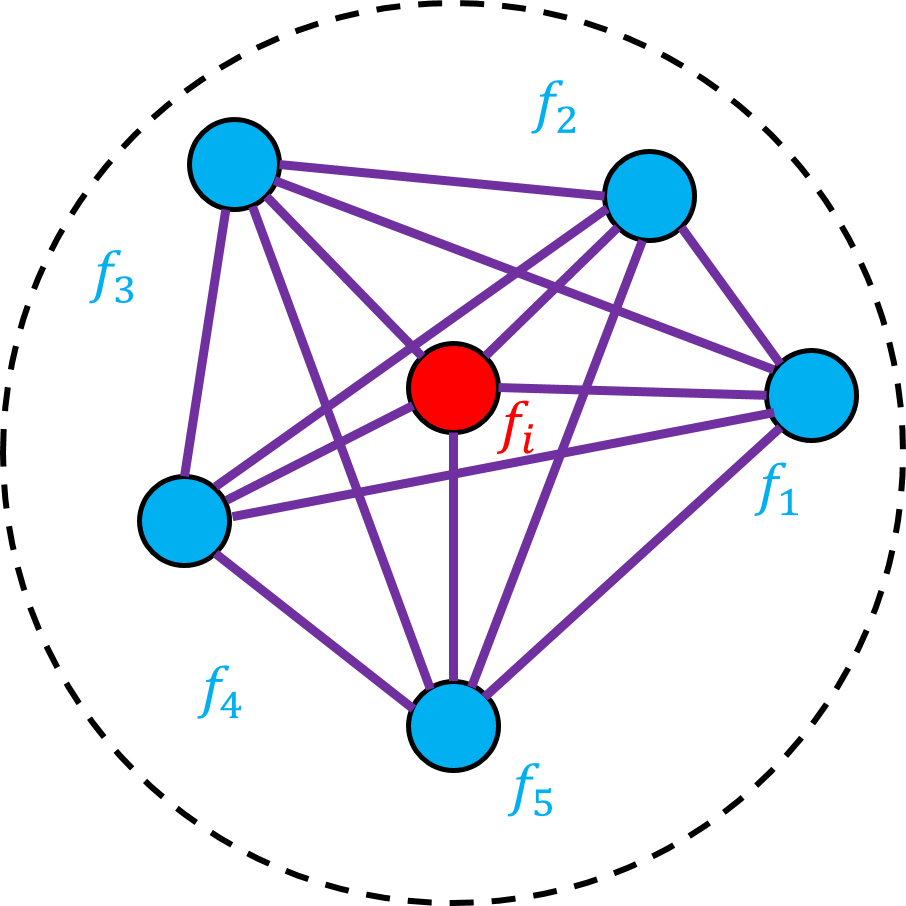}
    }
    \caption{Comparison of feature aggregation methods. (a) Features from the neighboring frames are weighted equally during aggregation. (b) The current aggregation methods only reason the relations between the current frame and neighboring frames. (c) Our proposed method computes every pair of frames in the neighborhood in the aggregation process.}
    \label{fig:problem}
\end{figure}

A primary challenge of video object detection is to tackle the feature degradation on video frames caused by camera jitter or fast motion. Under the circumstance, detection algorithms for still images are ill-posed for video tasks. Nonetheless, the video has rich temporal information, on which the same object may appear in multiple frames for a certain time span. The value of such temporal information is explored in prior studies using the post-processing paradigm \cite{han2016seq,kang2017t,kang2017t,lee2016multi}. These methods firstly perform still-image detection on single frames and then assemble the detection results across temporal dimensions using a disjoint post-processing step (i.e., motion estimation and object tracking). None of the above methods, therefore, operate in an end-to-end fashion. Moreover, if detection on single frames produces weak predictions, the assembling approach cannot improve the detection results.\\ 
\indent Alternatively, there have been several attempts to boost the performance of video detection using feature aggregation. \cite{liu2020video,voigtlaender2019mots,zhu2017deep} leverage optical flow to model the feature movement across frames and propagate temporal features to increase the feature representation for detection. With stronger features, the detection results are significantly improved. However, such temporal features are exploited by an intuitive lumping operation, which is oversimplified.\\ 
\indent In terms of how to organize features in aggregation, we recognize two important predecessors, FGFA \cite{zhu2017flow} and SELSA \cite{wu2019sequence}. Compared to the lumping solution \cite{liu2020video,voigtlaender2019mots,zhu2017deep}, both methods use similarity scores to select more helpful features for aggregation. The aggregated feature is organized by an adaptive weight at every spatial location for their representations
% This process can be visualized as Figure \ref{fig:problem}(a) where the red rectangle and dot represent the current frame and its corresponding feature and the blue rectangles and dots mean the neighboring frames and their related features 
% More specifically, these first perform weighted average on the related features from the neighboring frames, which are then aggregated into the feature of the current frame (as shown in Figure \ref{fig:problem}(a)).\\ 
% More specifically, these methodsfirst perform weighted average on the related features from the neighboring frames, which are then aggregated into the feature of the current frame
(as shown in Figure \ref{fig:problem}(a)). Albeit being superior over the prior efforts, FGFA \cite{zhu2017flow} and SELSA \cite{wu2019sequence} encounter several obstacles  from achieving optimal performance: 
% 1) local coherence between temporal features and foreground sample is frequently misaligned; 2) the aggregated feature representation is redundant because some salient features are repeatedly aggregated; 3) spatial positions for the region of interest on the temporal features are degraded after modeling their movements. 
1) They focus on modeling the global relation for every neighboring frame while ignoring the preservation of the local spatial information for aggregation; 
2) They primarily consider the global feature relations to the current frames, while having no constraint in feature learning among the neighboring frames (see Figure \ref{fig:problem}(b)); 3) They take a fixed number of neighboring frames for the feature aggregation, which is heuristic than general.  
% Thus, using adaptive weight for feature aggregation may not saturate the power of increasing feature representation for video object detection.\\

In this work, we attempt to take a deeper look at video object detection and improve the performance guarantees by organizing temporal information in a more rigorous principle. Inspired by \cite{zhu2017flow,wu2019sequence, chen20mega}, we propose TF-Blender to organically model features consistently and correspondingly in two ranges. Specifically, we reinforce local similarity in feature space on sequential video frames to depict the continuous and coherence of visual patterns, while identifying semantic correspondence across frames, which makes the temporal representations robust to appearance variations, shape deformations, and local occlusions. In this design, TF-Blender is able to generalize feature aggregation by encouraging the video representation and capturing helpful visual content to improve detection performance.
% We carry out extensive ablation studies to discover the xxx, xxx, xxx, and xxx. 
Concretely, we are able to achieve the following contributions:
% 1. all channel attention...to manage features...
% 1. all foreground
% 2. feature relation...f1-f2, f1, f2, 
% 2. 3D gives us more accurate spatial feature distribution....\\ 
% 3. self feature is learnable and aggregated feature loss... 
\begin{itemize}
    % \item We apply relation learning to manage temporal information in feature learning, which facilitates feature salience by organically modeling the foreground attention. Without changing the architecture, our approach can improve the accuracy of video detection methods  by more than $0.5\%$ in mAP on the ImageNet VID benchmark.
    % \item We devise a relation blender module, which depicts the temporal feature relations and selects helpful features in aggregation to increase the temporal-spatial feature representation across frames.  
    % \item We introduce attention losses to add constraints on features before aggregation to improve the final video object detection accuracy.
    % \item We introduce an attention supervision  strategy which encourages the trained algorithm to effectively learn the valuable features not only from the current frame but also from prior frames for the later detection step.  
    % \item  The proposed xxx is general and flexible. Without any modification, our method can improve the detection behavior on any video detection methods and achieve new state-of-the-art on ImageNet VID benchmark \cite{russakovsky2015imagenet}.
    \item We propose a framework called TF-Blender, which depicts the temporal feature relations and blends valuable neighboring features to increase the temporal-spatial feature representation across frames. 
    \item In TF-Blender, we devise a temporal relation module to manage temporal information and a feature adjustment module to add constraints in feature learning to preserve spatial information during feature aggregation. We, therefore, organize the feature learning between every pair of frames and aggregate features in the whole neighborhood (see Figure \ref{fig:problem}(c))
    % \item In Blender-Net, we introduce temporal relation   to manage temporal information in feature learning and aggregated feature   to add constraints on features before aggregation to improve the final video object detection accuracy.
    \item Our method is general and flexible, which can be crafted on any detection network. With our novel feature enhancement strategy, we can obtain an absolute gain of more than $0.7\%$ in mAP on the ImageNet VID benchmark and $1.5\%$ in mAP on YouTube-VIS benchmark for recent state-of-the-arts methods.
    % boost most of the state-of-the-arts methods can achieve an improvement of accuracy by $0.5\%$ in mAP on the ImageNet VID benchmark and $1.5\%$ in mAP on YouTube-VIS benchmark.
\end{itemize}
 \begin{figure*}[!hbt]
    \centering
    \includegraphics[width=17cm]{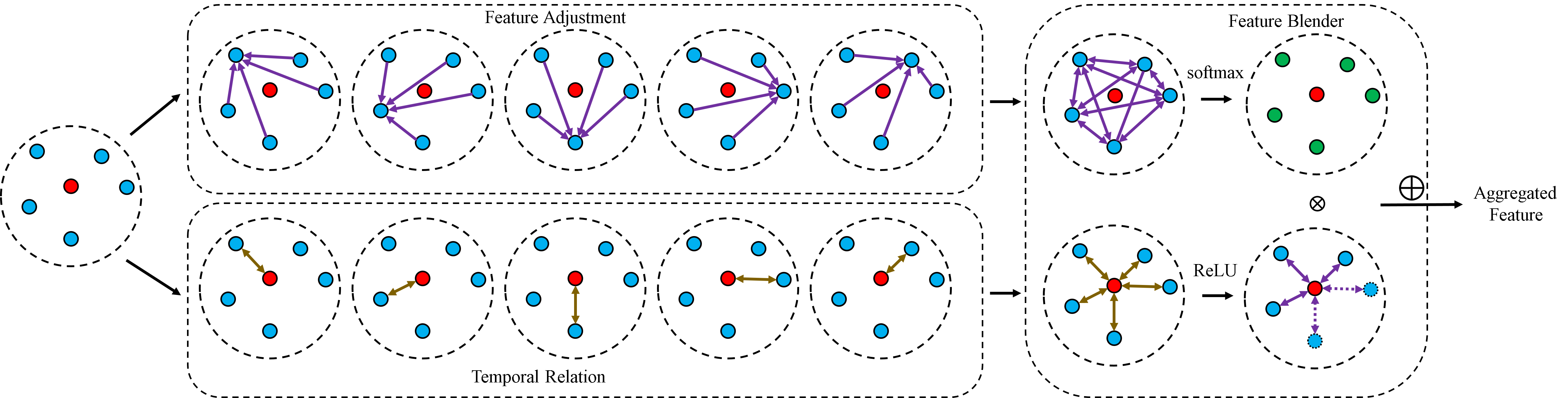}
    \caption{Our TF-Blender framework includes three key modules: a) \textbf{Temporal relation module}: Feature relation function $g\left(f_i,f_j\right)$ is used as input to learn adaptive weights $\mathcal{W}\left(f_i,f_j\right)$ used for feature blender. 2) \textbf{Feature adjustment module:} Every neighboring frame feature $f_j$ is aggregated with other neighboring features to generated feature representative $\mathcal{F}\left(f_i, f_j\right)$. 3) \textbf{Feature blender module:} The results of $\mathcal{W}\left(f_i, f_j\right)$ and $\mathcal{F}\left(f_i, f_j\right)$ are combined to aggregate the feature of the current frame with dynamic number of neighboring frames.}
    \label{fig:framework}
\end{figure*}
% feature prop--->feature aggregated...
% 1. importance of video task...\\
% 2. cannot use image detector...Current solution using temporal features\\
% 3. what's the problem: We need to address the problem. We can say propagate feature from flownet, but this approach has some issues, such as xxxx. To control the feature, people use adaptive weights. Can we say 3 limits about current methods? Can you please think maybe 2 limits about W??
% Feature management for propagation  can date back to xxx [23], a overly simplified assembling designs\\
\section{Related Works}
% \subsection{Image Object Detection}
% Conventionally, object detection focuses on image task. State-of-the-art methods \cite{liu2016ssd}\cite{huang2015densebox}\cite{ren2015faster}\cite{liu2016ssd} generally base on deep CNNs for feature extraction and then use a shallow structure for detection. Image detection can be divided by one-stage and two-stage approaches. One-stage methods (i.e., YOLO \cite{redmon2016you} and DenseBox\cite{huang2015densebox}) are generally inferior in accuracy compared to their two-stage counterparts (i.e., Faster R-CNN\cite{ren2015faster} and R-FCN \cite{dai2016r})  but can achieve a faster inference. In contrast to image object detection, we focus on the detection task on videos. Our method incorporates temporal information to increase the quality of convolutional feature maps, and consequently benefit the detection results.

\subsection{Video Object Detection}
\textbf{Video object detection.} Different from image object detection, video object detection faces challenging cases (i.e., motion blur, occlusion, and defocus) which rarely occur in images \cite{CORES2021104179, geng2020objectaware, 10.1145/3422852.3423477}. To handle the challenges in video domains, several works \cite{Kang_2016, kang2017t, han2016seq} use post-processing techniques on top of still image detectors. For instance, Seq-NMS \cite{han2016seq} links bounding boxes across frames with IoU threshold and re-rank the linked bounding boxes; TCN \cite{Kang_2016} introduces tubelet modules and applies a temporal convolutional network to embed temporal information to improve the detection across frames; T-CNN \cite{kang2017t} applies image object detectors to generate results and then uses optical flow to associate the detected results. Although achieving improvements,  none of them are trained end-to-end and their performances are still sub-optimal.

Another focus of the recent works \cite{zhu2017deep, zhu2017flow, wu2019sequence, deng2019relation, chen20mega, zhu2017high, xiao2018video} is to aggregate temporal features to improve the feature representation for detection. These methods can be divided into three categories: local aggregation, global aggregation, and combination aggregation. Local aggregation methods \cite{zhu2017flow, Wang_2018_ECCV, deng2019relation, liu2020video, zhu2017high, xiao2018video, feichtenhofer2018detect, bertasius2018object} usually focus on propagating features in a short range on video sequences. Among them, FGFA \cite{zhu2017flow} and MANet \cite{Wang_2018_ECCV} are representatives which use optical flow \cite{ilg2016flownet, fischer2015flownet} to calibrate and aggregate features across local frames. On the contrary, global aggregation methods \cite{wu2019sequence, 9010864, 9011008} rely on long-range semantic information. One seminal work is from SELSA \cite{wu2019sequence}, who computes the semantic similarity between the current frame and its neighbours across the whole video in order to perform temporal feature aggregation. Different from the methods which exploit features locally or globally, MEGA \cite{chen20mega} introduces a memory module to use both local and global features to enhance the visual representation of the current frame. The aggregation methods achieve further performance gain over the post-processing methods, but they generally focus on higher-level video frame selection instead of exploring lower-level temporal features exploitation.

\textbf{Video instance segmentation.} Similar to video object detection, MaskTrack R-CNN \cite{yang2019video} extends instance segmentation \cite{yolact-iccv2019, yolact-plus-tpami2020} from image domain to video domains which requires segmenting and tracking instances across frames. However, most of the current methods like MaskProp \cite{bertasius2020classifying}, EnsembleVIS \cite{Luiten_2019_ICCV} focus on how to track instances across frames rather than how to generate high-quality features for detection, segmentation, and tracking. In this work, we, therefore, propose a more principled solution, which effectively transforms and exploits valuable temporal features for the video object detection task. 
% \subsection{Video Instance Segmentation}
% Similar to video object detection, MaskTrack R-CNN \cite{yang2019video} extends instance segmentation from image domain to video domains. Besides detecting and segmenting instances in every frame, object tracking task is introduce to link the same instance across frames. MaskTrack R-CNN adds a tracking head to Mask R-CNN \cite{he2017mask} and treats object tracking as a multiple classification problem. Sipmask \cite{cao2020sipmask} divides instance segmentation mask into several sub-regions of a detected bounding box and applies every sub-region a spatial coefficient to improve segmentation mask quality.  All the methods mentioned above follow the ``detect then segment'' structure as Mask R-CNN \cite{he2017mask}. Therefore, segmentation performance is highly related to detection results. However, most of these current methods focus on how to track instances across frames or how to generate high-quality instance masks. 
\subsection{Relation Learning}
Relation learning is widely used for different tasks (i.e., point cloud analysis \cite{liu2019relationshape, CUI2021300} and image understanding \cite{liu2021densernet, yan2021hierarchical}) to describe the relationship between the current feature and its neighbors. RS-CNN \cite{liu2019relationshape} extends regular grid CNN to capture local point cloud features using geometric topology constraints among points. Similarly, PointConv \cite{wu2020pointconv} models the feature relation by computing both the local coordinates and point cloud density. Both methods capture local features in geometric space. On the contrary, DGCNN \cite{wang2019dynamic} defines EdgeConv which captures local point relation in high-dimensional feature space and updates the neighborhood for the kernel dynamically at each layer.\\ 
\indent Similarly, some recent works attempt to leverage relation learning for object detection. Inspired by \cite{hu2018relation} which proposes an object relation module for still image object detection, RDN \cite{deng2019relation} introduces a relation distillation network to aggregate features based on object relation to improving the features for video object detection. MEGA \cite{chen20mega} extends the relation learning from RDN and proposes a memory-enhanced global-local aggregation network, which organically manages long-range (global) features and short-range (local) features for aggregation in order to increase the feature representation of current time for detection. However, the focuses of the above methods \cite{hu2018relation,deng2019relation,chen20mega} are the selection of higher-level video frames for aggregation rather than modeling lower-level temporal relation to increasing the feature representation.\\
% model object features with a \textcolor{red}{non-learnable fashion}, which is sub-optimal to encode the temporal information. 
\indent Different from these methods, we propose a more general approach for relation learning in feature aggregation. Our TF-Blender can robustly depict the salient correspondences between the feature of the current frame and neighboring frames and exploit only valuable features for a stronger detection.

\section{TF-Blender}
% In this section, we start by introducing the overall pipeline of introducing the overall pipeline of introducing the overall pipeline of
% introducing the overall pipeline of

% the elaboration of each contribution module.
\subsection{Preliminary and Overall Pipeline}
% In this section, we elaborate on how to model the temporal relation across frames to improve feature aggregation for video object detection. 

The conventional feature aggregation methods~\cite{zhu2017flow,wu2019sequence, liu2020video, Wang_2018_ECCV} generally work in a constrained fashion. Given a set of neighboring frames $\textbf{F}_j$ of the current frame $\textbf{F}_i, \forall \textbf{F}_j \in \mathcal{N}\left(\textbf{F}_i\right)$, their corresponding features $f_j$ are weighted  equally based on the feature similarity to $\textbf{F}_i$ in order to aggregate the temporal feature $\Delta{f}_{i}$:
% Given a set of frames $\textbf{F}_j, \forall \textbf{F}_j \in \mathcal{N}\left(\textbf{F}_i\right)$ and their corresponding features $f_j$, we visualize the feature aggregation process of the current methods in Figure \ref{fig:currentMethod}. During feature aggregation process of the current methods, features $f_j$ from the nearby reference frame $\textbf{F}_j$ are weight averaged for the current key frame $\textbf{F}_i$ to get the aggregated feature $\Delta{f}_{i}$, as:
\begin{equation}
\begin{aligned}
    %   \hat{{f}_{i}} &= f_i + \Delta f_i\\
    %   &= f_i + \sum_{\textbf{F}_j \in \mathcal{N}\left(\textbf{F}_i\right) }({w}_{ij}\times{f}_{j}) \\
      \Delta f_i &= \sum_{\textbf{F}_j \in \mathcal{N}\left(\textbf{F}_i\right) }({w}_{ij}\times{f}_{j}).
\end{aligned}
\label{aggEq}
\end{equation}
The principal problem of feature aggregation, therefore, is to calculate weights $w_{ij}$ and select representative neighboring feature $f_j$. 
% Most of the concurrent works like FGFA \cite{zhu2017flow} and SELSA \cite{wu2019sequence} use cosine similarity or inner product between $f_i$ and $f_j$ as weight $w_{ij}$ and regard features $f_j$ as representative features for aggregation.
Different from the above simple paradigm, we exploit the temporal features from a general perspective. To achieve this goal, our TF-Blender crafts on three novel architectural modules, temporal relation module, feature adjustment module, and feature blender module, to boot the detection performances (see Figure \ref{fig:framework}).

\begin{figure}[!bt]
    \centering
    \subfigure[Input frames]{
    \includegraphics[width=8cm]{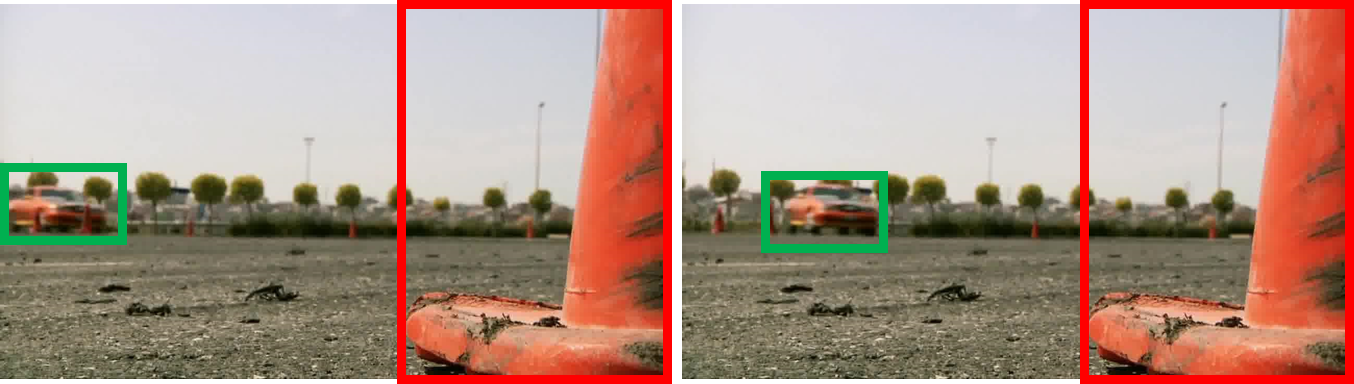}
    }
    \subfigure[Feature maps of input frames]{
    \includegraphics[width=8cm]{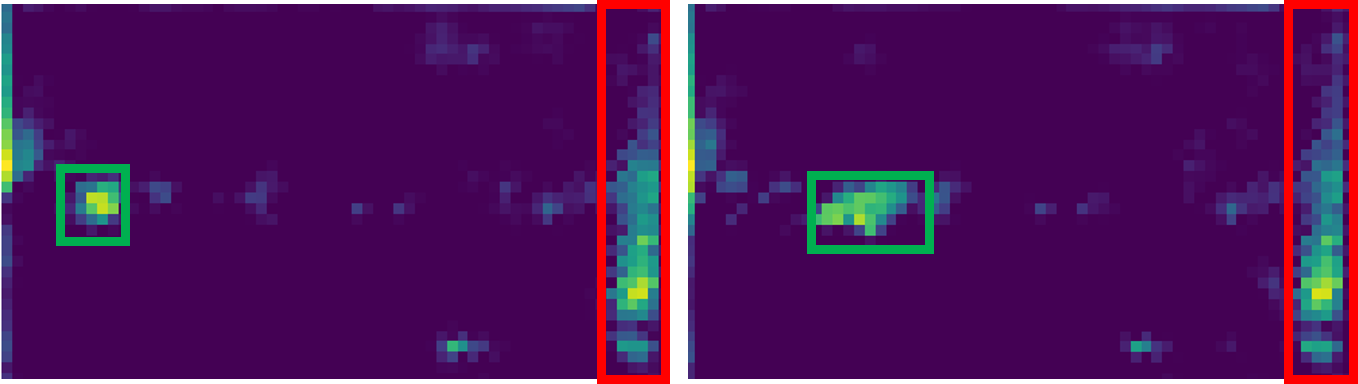}
    }
    \subfigure[Results of temporal relation]{
    \includegraphics[width=8cm]{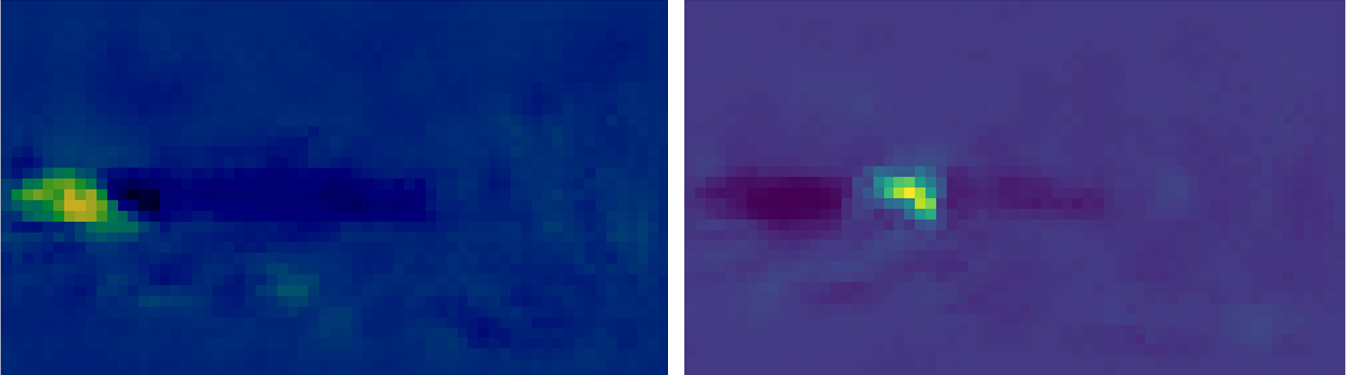}
    }
    % \subfigure[Results of temporal relation]{
    % \includegraphics[width=4cm]{img/inputAtt.png}
    % }
    \caption{An example of the problem of feature aggregation with global weights: a) shows two neighboring frames where the moving car (the green rectangles) is smaller than the traffic cone (the red rectangles). b) visualizes the feature maps of the two frames where the traffic cone also has a high response besides the car. With global weights, the high response feature of the traffic cone (the  red rectangles) cannot be suppressed unless the global weights have very small values. c) shows the results of our proposed temporal relation module which assigns every pixel in the feature map with an adaptive weight and can suppress the irrelevant features.}
    \label{fig:localGlobal}
\end{figure}
\subsection{Temporal Relation}
\begin{figure}
    \centering
    \includegraphics[width=5cm]{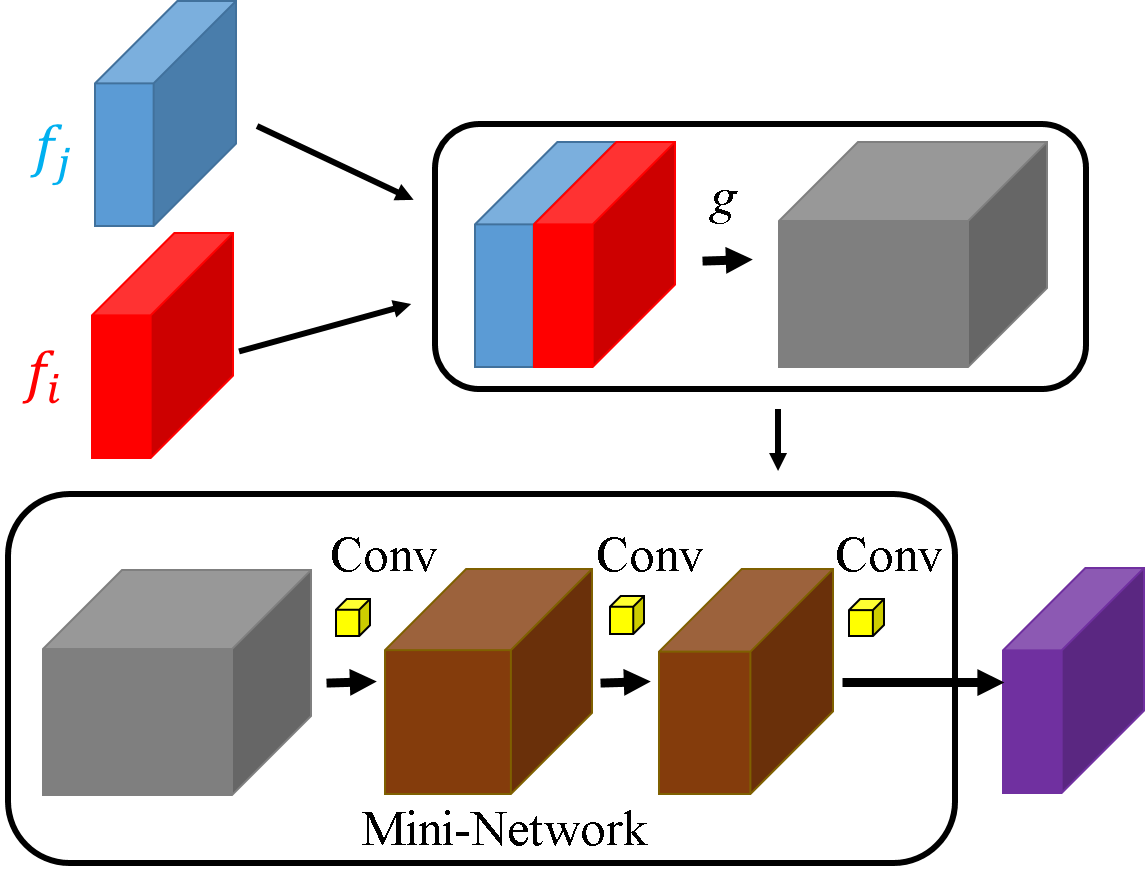}
    \caption{Visualization of temporal relation module. The input feature of $f_i$ and $f_j$ are visualized as blue and red cuboids respectively. Feature relation function $g$ models the temporal relation between $f_i$ and $f_j$ (the gray cuboids). In the mini-network, convolution layers (the yellow cubes) are applied to generate the final results (the purple cuboids). The results of the mid-layers are visualized as brown cuboids. }
    \label{fig:miniNetwork}
\end{figure}

% Our temporal relation module is designed to be a local weight with the same size as the representative features so that feature maps are weighted averaged pixel by pixel during aggregation process. We will discuss the problems of the global weights for aggregation in the current methods and introduce our proposed temporal relation.
Our temporal relation models the correspondences between the keyframe and its neighbors. To achieve this goal,
existing methods use $\mathbb{W}\left(f_i, f_j\right)$ to compute a global weight on every pixel in the feature map. This approach ignores local spatial information of the feature map during the process of aggregation, which causes the issue  of severe outliers in the feature map. 
% In this way, every pixel in the feature map cannot have its own weight during aggregation process. 
% Figure \ref{fig:localGlobal} shows an example where global weights have trouble with outliers in feature map. 
As shown in Figure \ref{fig:localGlobal}(a), two neighboring frames have a car with a fast speed and a still traffic cone marked with green and red rectangles respectively. 
% Due to the distances between the object and the camera, the moving car is in a small scale while the still traffic cone is in a large scale.
The feature maps of the input frames are visualized as Figure \ref{fig:localGlobal}(b) and the features of the traffic cone are outliers for the car detection. For global weights, if the weights between the paired frames are none-zero, irrelevant features cannot be removed during aggregation (see Figure \ref{fig:localGlobal}(b)). This problem occurs frequently when dealing with occlusions or small-scale objects.\\
\indent To address this issue, our temporal relation module generates adaptive weights $\mathcal{W}\left(f_i, f_j\right)$ for every pixel on the feature map in replace of the global weights $\mathbb{W}\left(f_i, f_j\right)$. 
We model $\mathcal{W}\left(f_i, f_j\right)$ as a tensor with the same size as the feature representatives for aggregation. For every neighboring frame $\textbf{F}_j$ of the current frame $\textbf{F}_i$, we use temporal relation module to calculate adaptive weights $\mathcal{W}\left(f_i, f_j\right)$ (see Figure \ref{fig:framework}). The process is formulated as:
\begin{equation}
    \mathcal{W}\left(f_i, f_j\right) = \mathcal{M}\left(g\left(f_i, f_j\right)\right),
    \label{temporalRelation}
\end{equation}
where $g$ is a feature relation function to describe the temporal relation between $f_i$ and $f_j$ and $\mathcal{M}$ is a masking function to calculate the adaptive weight based on $g$. As shown in Figure \ref{fig:localGlobal}(c), our temporal relation can enhance the feature representations from the region of interest and suppress the irrelevant features.\\  
% When $g$ is modeled as $f_i$ and $f_j$ themselves and $\mathcal{M}$ is cosine similarity function, Eq. (\ref{temporalRelation}) becomes:
% \begin{equation}
%     \mathcal{M}\left(g\left(f_i,f_j\right)\right) = \frac{f_i^T f_j}{|f_i||f_j|}
% \end{equation}
% which is widely used in the current methods.
% Instead of regarding $\mathcal{M}$ as simple function like cosine similarity, 
\indent More concretely, we compute $\mathcal{M}$ in Eq.~\ref{temporalRelation} using a mini-network (see Figure \ref{fig:miniNetwork}). Compared with the CoefNet in LMP \cite{10.1145/3422852.3423477}, our feature adjustment module is built on a lighter architecture, which makes our TF-Blender computationally efficient. The input of the module is $f_i$ and $f_j$, marked as red and blue cuboids respectively. Feature relation function $g$ describes the relation between $f_i$ and $f_j$ and generates the input (the gray cuboid) of the mini-network $\mathcal{M}$. Afterward, we apply three convolution layers (the yellow cubes) to generate the final adaptive weights $\mathcal{W}\left(f_i, f_j\right)$ (the purple cuboid). The selection of the feature relation function $g$ will be discussed in \ref{ID}.
% The purple cuboid is with the same size as the blue cuboid which serves as an attention mask for the input neighboring feature $f_j$.

% \begin{equation}
%     \Delta f_i = \sum_{\textbf{F}_j \in \mathcal{N}\left(\textbf{F}_j\right)} \left(\mathcal{W}\left(f_i, f_j\right) \cdot f_j\right)
% \end{equation}
\subsection{Feature Adjustment}
Our feature adjustment module aims to represent the feature consistency and salience of the neighboring frames for feature aggregation. A simple solution \cite{zhu2017flow, wu2019sequence, chen20mega} is to directly use feature $f_j$ from frame $\textbf{F}_j$ as the follow:
\begin{equation}
    \mathcal{F}\left(f_i,f_j\right) = f_{j\rightarrow i}.
\end{equation}
However, $f_j$ cannot be guaranteed to be valuable for aggregation as there is no constraints between these neighboring features. Therefore, we aggregate every neighboring frame feature $f_j$ before aggregating the current frame feature $f_i$. We get feature representative $\mathcal{F}\left(f_i, f_j\right)$ by aggregating $f_j$ with the other neighboring features $f_m, \forall \textbf{F}_m \in \mathcal{N}\left(\textbf{F}_i\right), \textbf{F}_m \neq \textbf{F}_j$ (see Figure \ref{fig:framework}). During feature adjustment, we use the temporal relation module to generate adaptive weights for neighbouring feature aggregation and the process can be expressed as:
\begin{equation}
    \mathcal{F}\left(f_i, f_j\right) = \sum_{\substack{\textbf{F}_m\in \mathcal{N}\left(\textbf{F}_i\right) \\ \textbf{F}_{m} \neq\textbf{F}_{j}}} \mathcal{W}\left(f_j,f_m\right)\otimes f_j 
\end{equation}
where $\otimes$ is element-wise multiplication, $f_m$ is the feature of the neighboring frame except itself, and $\mathcal{W}\left(f_j, f_m\right)$ is Eq~\ref{temporalRelation}, which can be expressed here as:
\begin{equation}
\begin{aligned}
\mathcal{W}\left(f_j, f_m\right) &= \mathcal{M}\left(g\left(f_j, f_m\right)\right) \\
\forall \textbf{F}_m &\in \mathcal{N}\left(\textbf{F}_i\right), \textbf{F}_m \neq \textbf{F}_j  
\end{aligned}
\end{equation}

% Therefore, our feature description module uses a different feature vector for $\mathcal{F}$ which is modeled as:
% \begin{equation}
%     \mathcal{F}\left(f_i,f_j\right) = \sum_{\substack{\textbf{F}_m\in \mathcal{N}\left(\textbf{F}_i\right) \\ \textbf{F}_{m} \neq\textbf{F}_{j}}}\left(w_m \left(f_j, f_m\right) \odot f_m\right)
% \end{equation}
% where $f_m$ is the neighboring feature of frame $\textbf{F}_i$ except frame $\textbf{F}_j$ and $w_m$ is learnable weights to aggregation.
% The second step is to blender the neighboring feature representatives from feature adjustment module with local weights from temporal relation module. We elaborate this process in the following section.

\begin{figure*} [!ht]
    \centering
    % \subfigure[Rare poses]{
    % \includegraphics[width=16cm]{img/compareEx2.png}
    % }
    % \subfigure[Part occlusion]{
    % \includegraphics[width=17cm]{img/compareEx3.png}
    % }
    % \subfigure[Fast motion]{
    % \includegraphics[width=17cm]{img/visExample.png}
    % }
    \includegraphics[width=17cm]{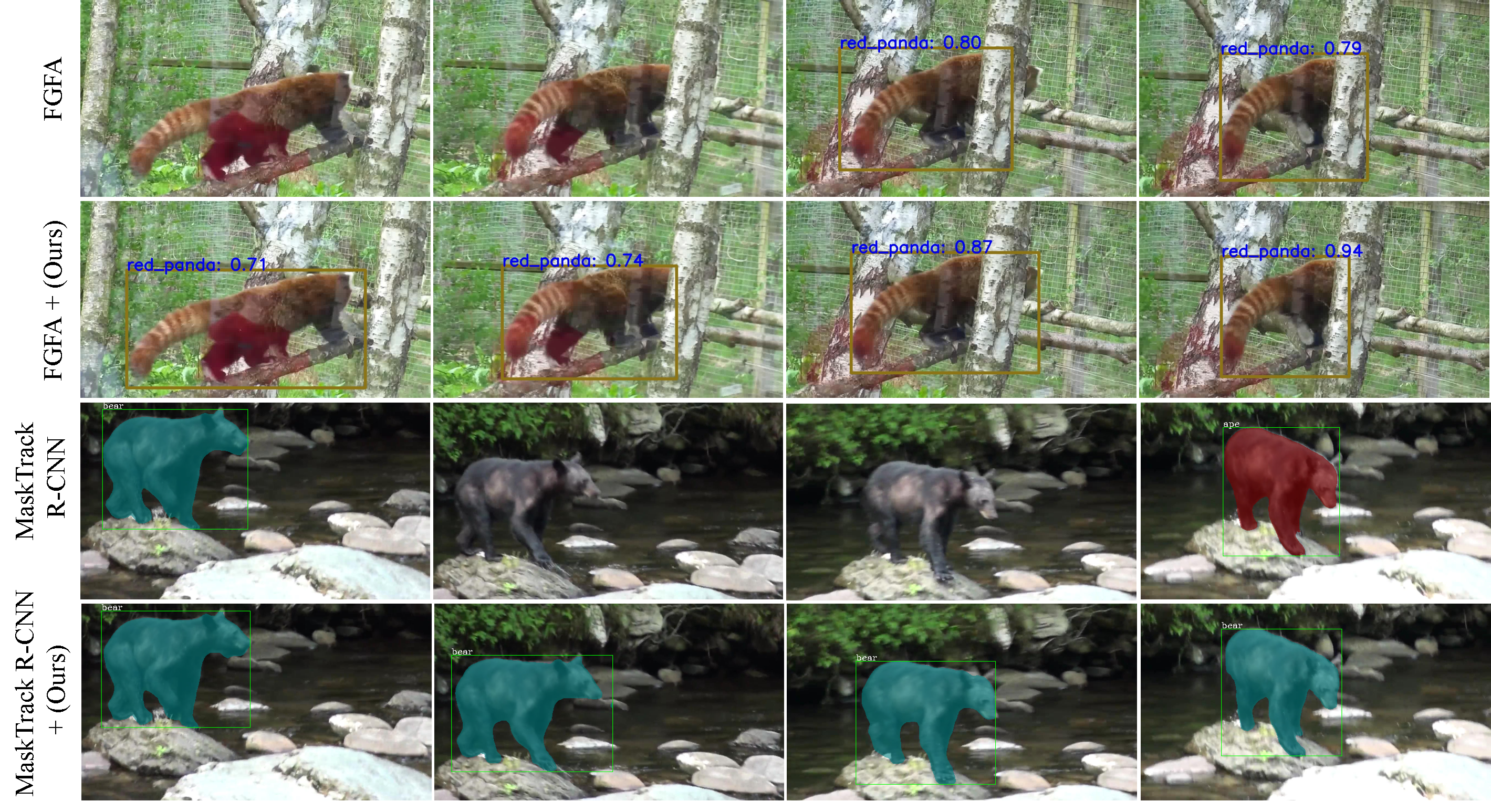}
    \caption{Quantitative examples of comparison between methods without and with our TF-Blender integrated on ImageNet VID and YouTube-VIS benchmarks.
    % Examples of detection results with other methods. (a) is an example of part occlusion situation on ImageNet VID benchmark. (b) is an example of rare pose situation on ImageNet VID benchmark. (c) is an example of fast motion situation on YouTube-VIS benchmark.
    }
    \label{fig:quatitiveResults}
\end{figure*}

\subsection{Feature Blender}
% After temporal relation module, we can get local weights to describe the relationship between $f_i$ and $f_j$, shown as the purple double arrows on the bottom of the right round rectangles in Figure \ref{fig:framework}. Feature adjustment module generates representative features of every neighboring frame for feature blender, shown as the green dots on the top of the right round rectangles in Figure \ref{fig:framework}. 

In our feature blender module, we first enhance the results of the temporal relation module with the non-linear function ReLU so that the contrast between the area of interests and background can be captured (see the blender module in Figure \ref{fig:framework}).  We formulate this process as:

\begin{equation}
    \hat{\mathcal{W}}\left(f_i, f_j\right) = \texttt{ReLU}\left(\mathcal{W}\left(f_i, f_j\right)\right).
\end{equation}
Meanwhile, we normalize the results of the feature adjustment module with the softmax function over all the channels to improve the generalization of our model. On the top of the feature blender module in Figure \ref{fig:framework}, blue dots are normalized to green dots by the softmax function with the guidance of purple double arrows.  The process can be expressed as:

\begin{equation}
    \hat{\mathcal{F}}\left(f_i, f_j\right) = \texttt{softmax}\left(\mathcal{F}\left(f_i, f_j\right)\right).
    \label{softmas}
\end{equation}

In our feature blender module, we force $\hat{\mathcal{W}}\left(f_i, f_j\right)$ to be $0$ if the adjusted neighboring feature is very similar to the feature of the current frame, shown as dashed purple double arrows in the feature blender module part of Figure \ref{fig:framework}. We use the cosine distance to describe the similarity between $\hat{\mathcal{F}}\left(f_i, f_j\right)$ and $f_i$. If the cosine distance is bigger than $\delta$, $\hat{\mathcal{W}}\left(f_i, f_j\right)$ is force to be $0$. We define this process as:

\begin{equation}
    \hat{\mathcal{W}}\left(f_i, f_j\right) = 0, \quad \textbf{if} \quad \frac{\hat{\mathcal{F}}\left(f_i, f_j\right)\cdot f_i}{|\hat{\mathcal{F}}\left(f_i, f_j\right)||f_i|} > \delta.
    \label{delta}
\end{equation}
We have this design because most of the current feature aggregation-based methods \cite{chen20mega, wu2019sequence, deng2019relation, zhu2017flow} have a fixed number of neighboring frames in aggregation. However, for neighboring frames which include issues of severe motion blur or defocus, aggregating them are irrelevant and redundant, which may cause unwanted ambiguity. 

Finally, we use element-wise multiplication to combine the results of from Eq.~\ref{softmas} and Eq.~\ref{delta} to perform the feature aggregation: 
\begin{equation}
\begin{aligned}
      \Delta f_i &= \sum_{\textbf{F}_j \in \mathcal{N}\left(\textbf{F}_i\right) }\left(\hat{\mathcal{W}}\left(f_i, f_j\right)\otimes \hat{\mathcal{F}}\left(f_i, f_j\right)\right) \\
    % \hat{\mathcal{W}}\left(f_i, f_j\right) &=  \texttt{ReLU}\left(\mathcal{W}\left(f_i, f_j\right)\right) \\
    % \hat{\mathcal{F}}\left(f_i, f_j\right) &= 
    % \sum_{\substack{\textbf{F}_m\in \mathcal{N}\left(\textbf{F}_i\right) \\ \textbf{F}_{m} \neq\textbf{F}_{j}}} \texttt{ReLU}\left(\left(\mathcal{W}\left(f_j,f_m\right)\right) \otimes \texttt{softmax}\left(f_j\right)\right) 
\end{aligned}
\end{equation}
% Then, $\Delta f_i$ will be the input of the further task networks for video object detection. We will discuss how to choose the parameters in our three modules in the following experiment section.
% Therefore, the whole process of our proposed method is summarized as Algorithm 1.
% \begin{algorithm}
%  \caption{}
%  \KwData{this text}
%  \KwResult{how to write algorithm with \LaTeX2e }
%  initialization\;
%  \While{not at end of this document}{
%   read current\;
%   \eIf{understand}{
%   go to next section\;
%   current section becomes this one\;
%   }{
%   go back to the beginning of current section\;
%   }
%  }
% \end{algorithm}

\begin{table}[!bt]
    \centering
    \begin{tabular}{c|c|c}
    \toprule
    Methods  & mAP(\%) & Runtime(FPS)\\
    \midrule
    % FGFA & \multirow{4}{*}{ResNet-50} & 74.7\\
    % SELSA & & 78.4\\
    % RDN & & 76.7\\
    % MEGA & & 77.3\\
    % \midrule
    % FGFA(Ours) & \multirow{4}{*}{ResNet-50} & 76.0$_{\uparrow1.3}$\\
    % SELSA(Ours) & & 79.4$_{\uparrow1.0}$\\
    % RDN(Ours) & & 77.2$_{\uparrow0.5}$\\
    % MEGA(Ours) & & 77.9$_{\uparrow0.6}$\\
    % \midrule
    % THP \cite{zhu2017high} & \multirow{3}{*}{ResNet-101 + DCN} & 78.6\\
    % STSN \cite{bertasius2018object} & & 78.9\\
    % OGEMN \cite{9011008} & & 80.0\\
    % \midrule
    FGFA\cite{zhu2017flow}  & 77.8 & 7.3\\
    SELSA\cite{wu2019sequence} & 81.5 & 10.6\\
    % MANet \cite{Wang_2018_ECCV} & & 78.1 \\
    RDN\cite{deng2019relation}  & 81.7 & -\\
    MEGA\cite{chen20mega} &  82.9 & 5.3 \\
    \midrule
    FGFA(Ours) & 79.3$_{\uparrow1.5}$ & 6.9\\
    SELSA(Ours) & 82.5$_{\uparrow1.0}$ & 10.1\\
    % MANet(Ours) & & 79.0$_{0.9}$\\
    RDN(Ours) & 82.4$_{\uparrow0.7}$ & -\\
    MEGA(Ours) & 83.8$_{\uparrow0.9}$ & 4.9\\
    \bottomrule
    \end{tabular}
    \caption{Performance comparison with the recent state-of-the-art video object detection models on ImageNet VID validation set. The backbone is ResNet-101 and runtime is tested on a single RTX 2080Ti GPU.}
    \label{tab:vod}
\end{table}
\section{Experiments}
\subsection{Implementation Details}\label{ID}
\textbf{Evaluation metrics.}
Following \cite{zhu2017deep,zhu2017flow}, we report all results on
using the mean average precision (mAP).\\
\indent \textbf{Video object detection setup.} We evaluate our methods with MEGA~\cite{chen20mega}, SELSA~\cite{wu2019sequence},FGFA~\cite{zhu2017flow}, and RDN \cite{deng2019relation}, the three state-of-the-art systems. We perform our training and evaluation  on the ImageNet VID benchmark \cite{russakovsky2015imagenet}, which contains 3,862 videos for training and 555 videos for validation. Following the widely used protocols in \cite{zhu2017flow, chen20mega, wu2019sequence}, we train our model on a combination of ImageNet VID and DET datasets. We implement our method mainly based on the source code of the original method. The whole network is trained on 8 RTX 2080Ti GPUs with SGD. During the training and inference process, each GPU holds on one set of images or frames. During the training process, the encoder parameters are frozen and an NMS of 0.5 IoU is adopted to suppress detection redundancy. 

\textbf{Video instance segmentation setup.} We also evaluate our proposed method with state-of-the-art MaskTrack R-CNN~\cite{lin2020video} and SipMask~\cite{cao2020sipmask}. We perform our training and evaluation  on the YouTube-VIS benchmark \cite{yang2019video}, where there are 3,471 videos for training and 507 videos for validation. During the training process, we use weights pretrained on MS-COCO \cite{lin2014microsoft} and use 8 RTX 6000 GPUs with SGD. In both training and evaluation, the original frame sizes are resized to $640 \times 360$.

\textbf{Parameters.} For mini-network $\mathcal{M}$ in Eq. (\ref{temporalRelation}), a three-layer CNNs is introduced to adapt the channels for feature aggregation. Feature relation function $g$ is defined as a concatenated tensor of $f_i$, $f_j$, $f_i - f_j, f_j - f_i$ and the $\delta$ in Eq. (\ref{delta}) is set to 0.7.

\subsection{Main Results}
\indent \textbf{Results on ImageNet VID benchmarks.} We compare state-of-the-art systems crafted on our method with their original implementations. For a fair comparison, we used the codes provided by the original papers and re-implement them with our proposed method. The results are demonstrated in Table \ref{tab:vod}. Based on the results, our proposed methods substantially improve the performance of every compared method listed in the table with the same backbone.\\
\indent For head-to-head comparisons, 
all the methods with the same backbone can leverage our proposed methods to improve their performances on detection results around $0.7\%$-$1.5\%$ on accuracy. Among them, FGFA with our proposed method has the highest improvement compared with other methods. Among them, local aggregation and global aggregation methods like FGFA \cite{zhu2017flow} and SELSA \cite{wu2019sequence} can have a better improvement with our proposed methods compared with combination aggregation methods like RDN \cite{deng2019relation} and MEGA \cite{chen20mega}. We argue that the limited performance gains come from the combination aggregation methods, which consider both local and global features and make detection more robust to issues like motion blur in videos.\\ 
% Despite of this, our proposed methods can still improve the performance of combination aggregation methods by more than $0.7\%$, which proves their effectiveness. 
\indent Figure \ref{fig:quatitiveResults} shows some examples of detection results with our methods integrated. Based on the examples, we can see that our proposed method can help solve the problem of weak detection with rare pose and part occlusion situations.
% More experiment results are shown in the supplementary materials. 

\begin{table*}[!bt]
    \centering
    \begin{tabular}{c|c|c|c|c|c|c|c}
    \toprule
    Methods & Category &  AP & AP$_{50}$ & AP$_{75}$ & AR$_1$ & AR$_{10}$ & FPS\\
    \midrule
    % DeepSORT \cite{wojke2017simple} & & 26.1 & 42.9 & 26.1 & 27.8 & 31.3 & -\\
    % FEELVOS \cite{voigtlaender2019feelvos} & & 26.9 & 42.0 & 29.7 & 29.9 & 33.4 & -\\
    % OSMN \cite{yang2018efficient} & & 27.5 & 45.1 & 29.1 & 28.6 & 33.1 & -\\
    Stem-Seg \cite{athar2020stemseg} & \multirow{6}{*}{One-stage} & 30.6 & 50.7 & 33.5 & 31.6 & 37.1 & 12.1 \\
    Stem-Seg(Ours) & & 31.3 & 51.5 & 34.1 & 32.1 & 37.9 & 11.3\\
    SipMask \cite{cao2020sipmask} & & 33.7  & 54.1 & 35.8 & 35.4 & 40.1 & 28.0\\
    SipMask(Ours)  &  & 35.1 & 55.5 & 36.9 & 36.1 & 41.3 & 26.6\\
    % SipMask \cite{cao2020sipmask} & & 33.7  & 54.1 & 35.8 & 35.4 & 40.1 & 28.0\\
    % SipMask(Ours)  &  & 35.1 & 55.5 & 36.9 & 36.1 & 41.3 & 26.6\\
    % SipMask++(Ours) \\
    SG-Net \cite{Liu_2021_CVPR} & & 34.8 & 56.1 & 36.8 & 35.8 & 40.8 & 22.9\\
    SG-Net(Ours) & & 35.7 & 57.1 & 37.6 & 36.6 & 42.0 & 21.3\\
    % SipMask++ \\
    \midrule
    MaskTrack R-CNN \cite{lin2020video} & \multirow{2}{*}{Two-stage} & 30.3 & 51.1 & 32.6 & 31.0 & 35.5 & 10.0\\
    MaskTrack R-CNN(Ours) & & 31.4 & 52.3 & 33.5 & 31.9 & 36.5 & 9.4\\

    \bottomrule
    \end{tabular}
    \caption{Performance comparison with the recent state-of-the-art video instance segmentation models on YouTube-VIS validation set. The backbone is ResNet-50-FPN and the models are pretrained on MS-COCO. The runtime is tested on a single RTX TITAN GPU.}
    \label{tab:vis}
\end{table*}

\textbf{Experiments on YouTube-VIS benchmark.} We also evaluate our proposed method on YouTube-VIS dataset \cite{yang2019video} and report our results on the validation as \cite{yang2019video, cao2020sipmask, athar2020stemseg}. Most of the current video instance segmentation methods focus on how to generate high-quality masks and link the same objects across frames with features extracted by the backbones like ResNet while only a few of them pay attention to improve the features for mask generation and object tracking. We add our proposed methods to these video instance segmentation methods to evaluate the effectiveness of our TF-Blender on issues like motion blur and defocus in videos. The results with ResNet-50 as backbones are shown in Table \ref{tab:vis}.
% and those with ResNet-101 are shown in the supplementary materials.
From Table \ref{tab:vis}, our proposed methods achieve competitive results under all evaluation metrics. With our proposed methods, MaskTrack R-CNN and SipMask can be improved by more than $1.6\%$ on the AP metric. The bottom part of Figure \ref{fig:quatitiveResults} shows an example of detection and segmentation results with our integrated.
\begin{table}[!bt]
    \centering
    \begin{tabular}{c|c|c|c|c}
    \toprule
    Method & TR & FA & FB &mAP(\%) \\
    \midrule
    a & & & & 77.8 \\
    b & \checkmark & & & 78.5\\
    c & & \checkmark & & 78.1\\
    d & & & \checkmark & 78.3\\
    e & \checkmark & \checkmark & & 78.6\\
    f & \checkmark & & \checkmark & 78.8\\
    g & & \checkmark & \checkmark & 78.5\\
    h & \checkmark & \checkmark & \checkmark & 79.3\\
    % \midrule
    % TR & & \checkmark & & \checkmark &  \checkmark\\
    % FA & & & \checkmark & & \checkmark  \\
    % % BN & & & & \\
    % FB & & & & \checkmark & \checkmark  \\
    % \midrule
    % mAP(\%) & 77.9 & 78.7 & 78.5 & 79.1 & 79.3\\
    \bottomrule
    \end{tabular}
    \caption{Impact of integrating every functional module into the baseline to the accuracy. TR, FA, and FB stand for temporal relation module, feature adjustment module, and feature blender module respectively.}
    \label{tab:ablationStudy}
\end{table}
\subsection{Ablation Study}
We carry out extensive ablation studies to discover the optimal settings related to different settings of our system using FGFA \cite{zhu2017flow}.

\textbf{Analysis of contributing components.} We first conduct experiments on the effect of every component in our proposed method and the results are shown in Table \ref{tab:ablationStudy}. The baseline model a is the original FGFA. Every component of our proposed method (temporal relation, feature adjustment, and feature blender) contributes towards improving the overall performance in detection accuracy. By introducing the temporal relation module, the performance of model b can be improved by $0.7\%$. Model c adds our feature adjustment module to the baseline and gets an improvement of $0.3\%$ compared with the baseline model a. We add our feature blender module to model a to generate dynamic numbers of neighboring frames for feature aggregation and get model d, which is $0.5\%$ better than the original model on mAP metric. Model e, f, and g come from the combination of models a, b and c. As can be shown in Table \ref{tab:ablationStudy}, by combining every two of our proposed methods, the video object detection performance can be further improved. Compared with the baseline model a, our full model h can obtain an absolute gain of $1.5\%$ in accuracy of video object detection. 

\textbf{Analysis of temporal relation.}
We conduct ablation studies on the choice of $g$ in Eq. (\ref{temporalRelation}). During these experiments, all the other experimental settings are kept the same. We first try different combinations of $f_i$ and $f_j$ for $g$ on FGFA \cite{zhu2017flow} as Table \ref{tab:ablationStudyRB}. A naive idea is to use just $f_i$ and $f_j$ as input and there is $0.5\%$ improvement on FGFA. We think that the performance is limited because only individual frame features are taken into account which is not enough to describe the relationship between the $f_i$ and $f_j$. Thus, we introduce the difference between $f_i$ and $f_j$ to $g$ and get an improvement of $0.8\%$ for FGFA. We then use the summation of $f_i$ and $f_j$ as $g$ to generate $\mathcal{W}\left(f_i, f_j\right)$ but there is only $0.1\%$ improvement. We also make a combination between $f_i + f_j$ with the other choices mentioned above (like $f_i, f_j$, and $f_i - f_j$), but the results of the combination are worse than those of the original. We think the reason why $f_i + f_j$ is not suitable to describe the relations between $f_i$ and $f_j$ is $f_i + f_j$ works like an average filter which mixes the pixels with higher responses and those with lower responses in the feature map. Besides the experiments mentioned above, we also try $f_i, f_j, f_i - f_j$ and get an improvement of $1.1\%$. Finally, we choose $f_i, f_j, f_i - f_j, f_j - f_i$ as our feature relation function $g$, which has the highest detection accuracy.  Since $f_i$  and $f_j$ denote the current and adjacent features respectively. Frame $\textbf{F}_j$ could be a frame before or after the current frame $\textbf{F}_i$. Thus, it is imperative to calculate both $f_i - f_j$ and $f_j - f_i$, as they model the different temporal correspondence and consistency. 
\begin{table}[!bt]
    \centering
    \begin{tabular}{c|c}
    \toprule
    $g$ & mAP(\%)\\
    \midrule
    $f_i, f_j$ & 78.3\\
    $f_i - f_j$ & 78.6\\
    $f_i + f_j$ & 77.9\\
    $f_i, f_j, f_i + f_j$ & 78.1\\
    $f_i - f_j, f_i + f_j$ & 78.5\\
    $f_i, f_j, f_i - f_j$ & 78.9\\
    $f_i, f_j, f_i - f_j, f_j - f_i$ & 79.3\\
    % $f_i, f_j, f_i - f_j, f_i + f_j$ & \\
    \bottomrule
    \end{tabular}
    \caption{Results of different designs on feature relation function $g$.}
    \label{tab:ablationStudyRB}
\end{table}
% \indent\textbf{Analysis of feature blender.} We explore the effect of the threshold $\delta$ in our feature blender module for the aggregation process in Eq. (\ref{delta}). We use FGFA \cite{zhu2017flow} as the model for this experiment and keep all the other parameters fixed. We choose a range of $\delta$ from $0.1$ to $0.9$ and the experiment results are visualized as Figure \ref{fig:delta}. If $\delta$ is too small, the performance of video object detection even gets worse since too many weights are forced to be $0$ and those features with non-zero weights are too different from the feature of the current frame. On the contrary, if $\delta$ is too big which means most of the features from the neighboring frames are kept for aggregation, the performance is also not good enough. We think this is because these features from the neighboring frames are redundant or even misleading for the frames which do not have object detection issues caused by motion blur, defocus, and so on.
% \begin{figure}[!bt]
%     \centering
%     \includegraphics[width=5.5cm]{img/delta.png}
%     \caption{Improvement of performance with different $\delta$.}
%     \label{fig:delta}
% \end{figure}

\textbf{Experiments on $\mathcal{M}$.} We conduct experiments on the design of $\mathcal{M}$ for the temporal relation module, especially on the number of layers of $\mathcal{M}$ for the mini-network. Model a is the simplest design where there is only one convolution layer with kernel size $1\times1$. By keeping the kernel size fixed and adding one more convolution layer, model b can increase the mAP by $0.2\%$. When there are three convolution layers with kernel size $1\times1$, the detection accuracy can 
obtain $79.2\%$ as model c. However, when adding more convolution layers, as in model d, the detection accuracy begins to decrease. We argue that the increasing number of convolution layers introduces arduous parameters in the mini-network which cause overfitting. In model e, we change the kernel size from $1\times1$ to $3\times3$ and get an improvement of detection accuracy by $0.1\%$.

\begin{figure}[!bt]
    \centering
    \subfigure[Motion speed]{
    \includegraphics[width=3.8cm]{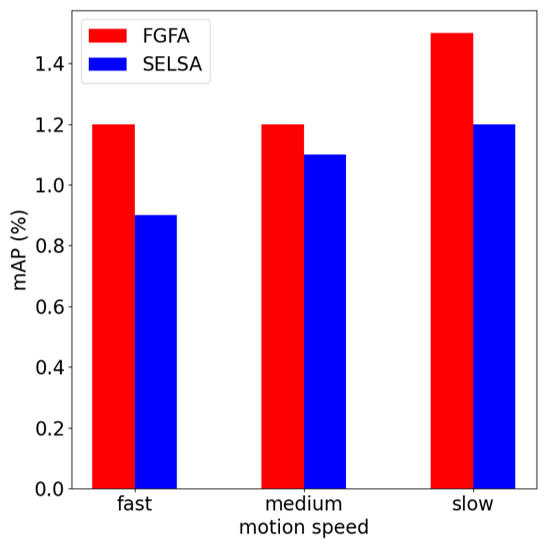}
    }
    \subfigure[Object sizes]{
    \includegraphics[width=3.8cm]{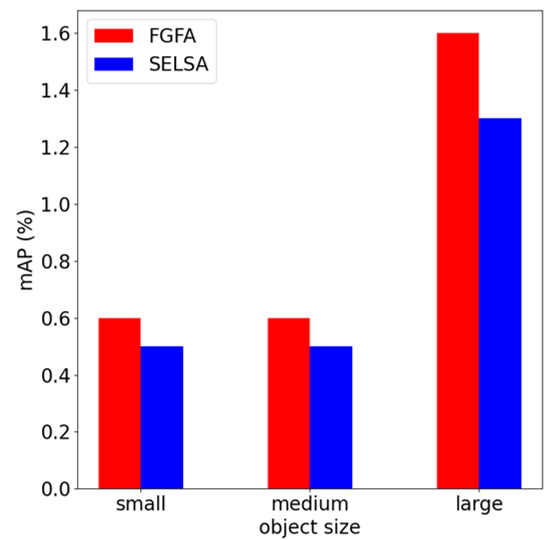}}
    \caption{Improvement of performance with different motion speeds and object sizes.}
    \label{fig:scaleAndSpeed}
\end{figure}

\begin{table}[!ht]
    \centering
    \begin{tabular}{c|c|c}
    \toprule
       model & \# of layers &  mAP(\%)\\
     \midrule
       a & 1 & 78.8\\
       b &  2 & 79.0\\
       c &  3 & 79.2\\
       d & 4 & 79.1\\
       e & 3 & 79.3 \\
     \bottomrule
    \end{tabular}
    \caption{Impact of the number of layers for $\mathcal{M}$.}
    \label{tab:mAblation}
\end{table}
\textbf{Analysis of object sizes and motion speeds.} We also investigate the effect of our TF-Blender on the object sizes and motion speeds of the objects. We use the same definition as MS-COCO \cite{lin2014microsoft} and FGFA \cite{zhu2017flow} for object sizes and motion speeds respectively. We use mAP as evaluation metrics and visualize the improvement of performance on objects with different sizes and motion speeds as Figure \ref{fig:scaleAndSpeed}.
We notice that our method has different improvements on objects with various motion speeds. As shown in Figure \ref{fig:scaleAndSpeed} (a), there is a higher improvement for objects with slow motion speeds compared with those with fast and medium speeds. We think that there may be two reasons. One is that even though our proposed method can help improve the detection accuracy for objects with fast motion speeds, it's still a challenge to have accurate enough detection results for all the objects with fast-motion speed. Another reason is that objects with slow-motion account for $37.9\%$ in ImageNet VID benchmark while those with medium and fast motion speeds are $35.9\%$ and $26.2\%$ respectively.

Another critical observation from our experiment that 
our method can offer the highest improvement for detection on large objects, as shown in Figure \ref{fig:scaleAndSpeed}(b). This resonates with the assumption of our proposed method: since large objects have larger feature map sizes, the corresponding pixel can benefit more from an individual weight for fine-grained feature encoding. For small objects, since their feature maps are small, the weights for aggregation have less contribution to feature representation improvement.

\textbf{Speed-accuracy tradeoff.} The computational loads for convectional methods (i.e., FGFA~[\textcolor{green}{47}] and SELSA~[\textcolor{green}{41}]) stem from two major sources: 1. feature extraction (encoding) network  $\mathcal{N}_{ex}$; 2. task network  $\mathcal{N}_{tk}$. Thus, the runtime complexity for the above methods is:
\begin{equation}
\begin{aligned}
    % \mathcal{O}\big(\cdot\big) =
    \mathcal{O}\big(\mathcal{N}_{ex}\big)+\mathcal{O}\big(\mathcal{N}_{tk}\big)
\label{11}
\end{aligned}
\end{equation}
\indent While the proposed TF-Blender approach is adopted, the computational cost can be defined as:
\begin{equation}
\begin{aligned}
    % \Hat{\mathcal{O}}\big(\cdot\big) =
    \mathcal{O}\big(\mathcal{N}_{ex}\big)+ i\cdot\mathcal{O}\big(\mathcal{N}_{tf}\big)+\mathcal{O}\big(\mathcal{N}_{tk}\big)
% \label{11}
\end{aligned}
\end{equation}
where $\mathcal{N}_{tf}$ is the cost for the TF-Blender module and $i$ is the number of aggregated frames.
% and $i$ is the number of frames. 
Typically, $\mathcal{O}\big(\mathcal{N}_{tk}\big) \ll \mathcal{O}\big(\mathcal{N}_{ex}\big)$ and $\mathcal{O}\big(\mathcal{N}_{tf}\big) \ll \mathcal{O}\big(\mathcal{N}_{ex}\big)$. Thus, the cost ratio $r$ can be expressed as:
\begin{equation}
\begin{aligned}
    r=1+\frac{i\cdot\mathcal{O}\big(\mathcal{N}_{tf}\big)}
{\mathcal{O}\big(\mathcal{N}_{ex}\big)+\mathcal{O}\big(\mathcal{N}_{tk}\big)}\\
\end{aligned}
    \label{13}
\end{equation}
This increasing computational cost is affordable because the impact of $i\cdot\mathcal{O}\big(\mathcal{N}_{tf}\big)$ is negligible. 

\begin{figure}[!ht]
    \centering
    \includegraphics[width=8cm]{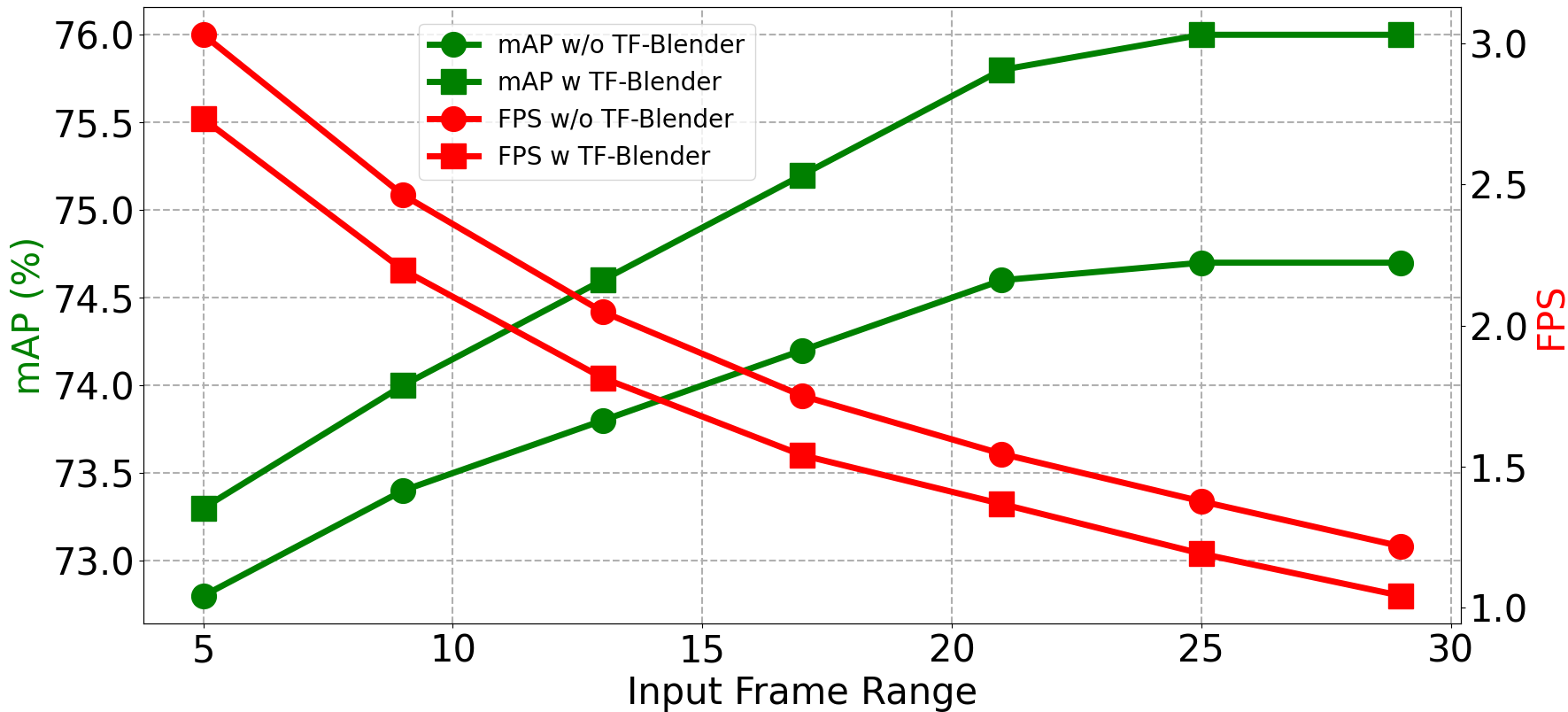}
    \caption{Demonstration of a speed-accuracy tradeoff with and without TF-Blender on FGFA with ResNet-50.}
    \label{fig:speed_acc_tradeoff}
\end{figure}

We visualize the speed-accuracy tradeoff of FGFA~[\textcolor{green}{47}] as an example (cf. Figure \ref{fig:speed_acc_tradeoff}). With the increasing number of input frames, FGFA with TF-Blender achieves significant improvement in accuracy while the runtime increase keeps in an affordable range.

\section{Conclusion}
In this paper, we discuss the problems of video object detection and introduce a framework named TF-Blender which contains temporal relation, feature adjustment, and feature blender modules to solve the problem of feature degrading in the video frames. Our method is flexible and general, which can be adopted by any learning-based detection network to achieve improved performance. Extensive experiments demonstrate that, with the integration of our proposed method, the current state-of-the-art methods can improve video object detection accuracy on ImageNet VID and YouTube-VIS benchmarks by a large margin. We believe that our TF-Blender can be a valuable addition to the existing methods for temporal feature  aggregation for video detection and TF-Blender can be extended to other video analysis tasks like video instance segmentation.
{\small
\bibliographystyle{ieee_fullname}
\bibliography{main}
}

\end{document}